\documentclass[a4paper,fleqn]{cas}

\usepackage[numbers]{natbib}
\usepackage{hyperref}
\usepackage{xcolor}
\hypersetup{colorlinks=true, citecolor=blue, linkcolor=blue, urlcolor=blue}
\def\tsc#1{\csdef{#1}{\textsc{\lowercase{#1}}\xspace}}
\tsc{WGM}
\tsc{QE}
\begin{document}
\let\WriteBookmarks\relax
\def\floatpagepagefraction{1}
\def\textpagefraction{.001}
\let\printorcid\relax % 可去掉页面下方的ORCID(s)

% \shorttitle{<short title of the paper for running head>} 
\shorttitle{Depth Awakens: A Depth-perceptual Attention Fusion Network for RGB-D Camouflaged Object Detection}    

% \shortauthors{<short author list for running head>}
\shortauthors{Xinran Liu et al.}

\title[mode = title]{Depth Awakens: A Depth-perceptual Attention Fusion Network for RGB-D Camouflaged Object Detection}

\author[1]{Xinran Liu}
\fnmark[1] 
\ead{lxr7766@stu.ouc.edu.cn} 

\author[1]{Lin Qi}
\fnmark[2] 
\cormark[1]
\ead{qilin@ouc.edu.cn}

\author[1]{Yuxuan Song}
\fnmark[3] 
\ead{syxvision@stu.ouc.edu.cn}

\author[1]{Qi Wen}
\fnmark[4] 
\ead{wenqi@ouc.edu.cn}

\address[1]{Department of Computer Science and Technology, Ocean University of China, Qingdao 266100, China}

\cortext[1]{Corresponding author} 
%TC:ignore
\begin{abstract}
Camouflaged object detection (COD) presents a persistent challenge in accurately identifying objects that seamlessly blend into their surroundings. However, most existing COD models overlook the fact that visual systems operate within a genuine 3D environment. The scene depth inherent in a single 2D image provides rich spatial clues that can assist in the detection of camouflaged objects. Therefore, we propose a novel depth-perception attention fusion network that leverages the depth map as an auxiliary input to enhance the network's ability to perceive 3D information, which is typically challenging for the human eye to discern from 2D images.
The network uses a trident-branch encoder to extract chromatic and depth information and their communications. Recognizing that certain regions of a depth map may not effectively highlight the camouflaged object, we introduce a depth-weighted cross-attention fusion module to dynamically adjust the fusion weights on depth and RGB feature maps. To keep the model simple without compromising effectiveness, we design a straightforward feature aggregation decoder that adaptively fuses the enhanced aggregated features. Experiments demonstrate the significant superiority of our proposed method over other states of the arts, which further validates the contribution of depth information in camouflaged object detection. The code will be available at https://github.com/xinran-liu00/DAF-Net.
\end{abstract}


% Keywords
%TC:endignore
\maketitle

% Main text
\section{Introduction}
Camouflaged object detection (COD) is a challenging task in the field of computer vision. The goal of COD is to accurately segment camouflaged objects from their surroundings. Beyond traditional applications in biology, COD has become increasingly prevalent in various tasks, such as polyp segmentation\cite{fan2020pranet}, defect detection\cite{zeng2022small} and pest detection\cite{cheng2017pest} in recent years.

The advancement of deep learning has accelerated the research in COD. Various deep learning-based approaches have been proposed, with some of them focusing on incorporating auxiliary cues such as edge\cite{jia2022segment}, texture\cite{ren2021deep, zhu2021inferring} and frequency domain information\cite{zhong2022detecting} to improve detection accuracy.

\begin{figure}[h]
\centerline{\includegraphics[width=\columnwidth]{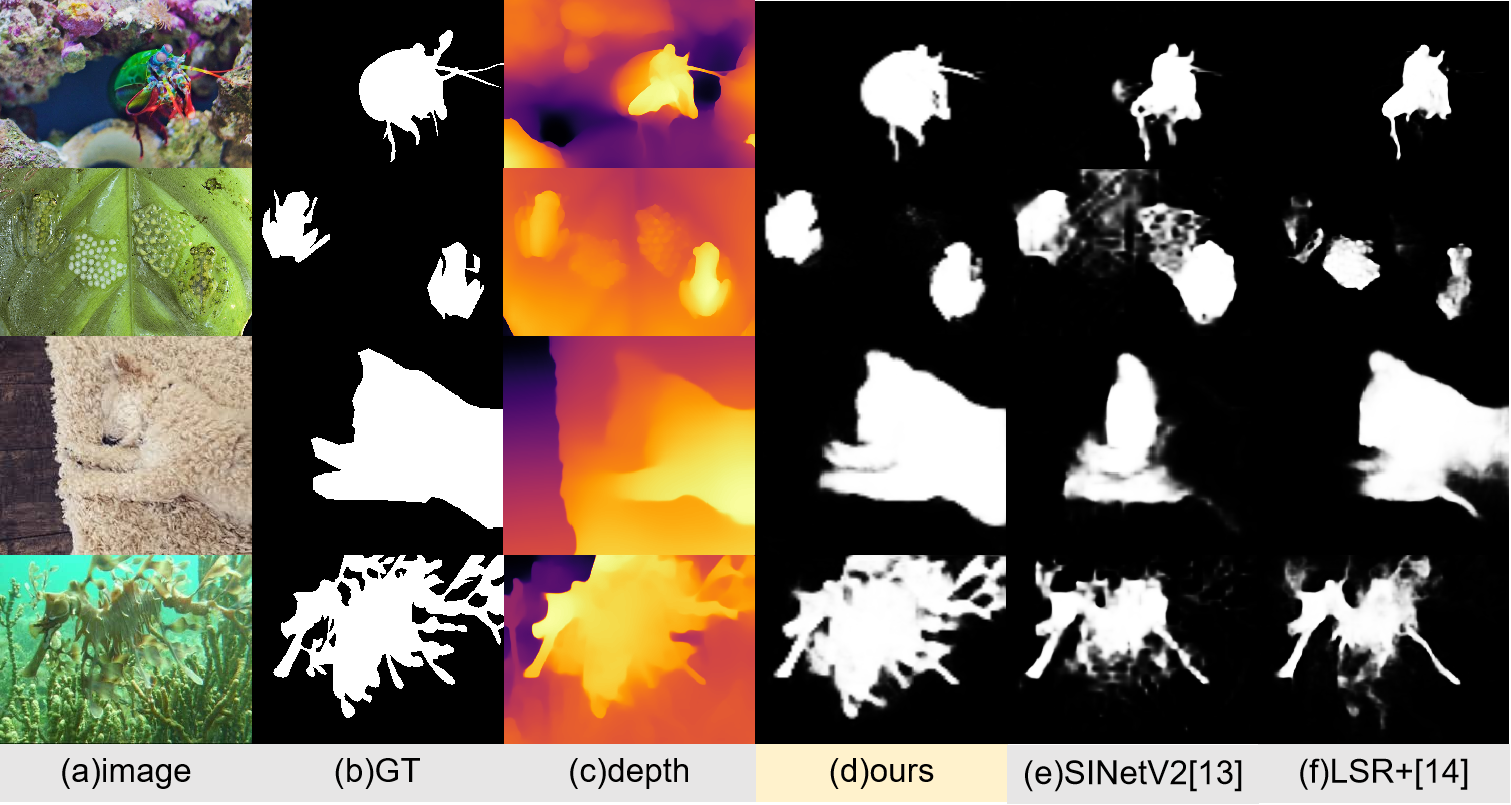}}
\caption{Visual examples of different methods. (a) RGB images. (b) Ground truths. (c) Depth maps. (d) Our results. (e)-(f) Prediction maps produced by SINet-V2\cite{fan2021concealed} and LSR+\cite{lv2023towards}, respectively.}
\label{fig:fig1} 
\end{figure}

Despite the recent advancements in COD, current methods still face difficulties in achieving complete object segmentation in challenging scenarios such as complex environments and situations where objects have textures that closely resemble their surroundings. Figure \ref{fig:fig1} illustrates some failures of existing methods.
This indicates that chromatic cues such as the utilization of edges and textures may reach its ceiling and cannot provide further improvement in COD tasks.

While camouflage assists animals in their survival and hunting, it primarily relies on 2D appearance rather than 3D geometry.
Since humans and animals operate within a genuine 3D context and are able to perceive distance to objects, spatial cues like depth, shape, surface normal and internal consistency can greatly aid in the detection process, as illustrated in Figure \ref{fig:fig1}(c).
The objects that are originally concealed in chromatic images will, in contrast, ``pop out'' in the depth images.
This transformation shifts the perspective from camouflaged object detection to salient object detection.

The community of salient object detection (SOD) \cite{chen2018progressively, zhao2019contrast, li2020asif} has made remarkable advancements by leveraging both RGB and depth cues, whereas there have been only very few attempts in COD utilizing depth as supervision\cite{xiang2021exploring, wu2023source}. 
% The pioneer work described in \cite{xiang2021exploring} has discovered the contribution of depth in COD task as an auxiliary branch but fails to fully exploit the potentials embedded in depth maps. Additionally, the lack of rigorous evaluation criteria to constrain the quality of depth maps consequently has a detrimental effect on the accuracy of their prediction results.
How to exploit depth cues for COD is still under exploration.
Another limitation in RGB-D COD research is that there is no COD dataset with depth ground-truth. Most COD datasets are collected from the Internet where depth is unavailable in most practical scenarios when capturing, which leads researchers to the indirect way of obtaining depth data in COD images -- Single Image Depth Estimation (SIDE).

Inferring scene depth from a monocular image has been a hot topic in the field of computer vision.  Despite the existence of several effective methodologies, the presence of a one-to-many mapping (from a 3D scene to a 2D image) renders this task inherently ill-posed. Most solutions to this problem are inspired by how humans perceive scene depth (i.e. prior knowledge learned from the real world), where depth estimators can be learned from extensive datasets captured with depth sensors, allowing these models to effectively generalize to unseen content.
It is noteworthy that within this process, the model predominantly attends to 
elementary cues related to geometric structure, such as photometric continuity, ground connections, vanishing points and object edges \cite{mertan2022single}. 
The model's capacity to acquire an extensive repertoire of low-level features endows it with the capability to proficiently estimate depth, even when directly confronted with a COD task that diverges from its original semantic context.
This observation motivates us to employ the knowledge acquired from these models as a prior in the context of the COD task.

% 讲述别人的工作，过去式？一般式？
We have investigated the work of \cite{xiang2021exploring} on the contribution of generated depth maps in COD, where the authors preliminarily fused the depth map as input but found the generated depth leads to inferior performance in the COD task. In their approach, depth maps are not used as input but instead are employed as supervision for an auxiliary branch. 
In light of this, we pose two inquiries:  1) Can the depth map only be used as a supervision to fully exploit its effective information? 2) Is there any straightforward fusion approach to combine the generated depth maps and RGB maps reasonably?
The previous work \cite{xiang2021exploring} and our experiments have found that only very few generated depth maps exhibit some errors in some parts of the object, whereas in most cases the 3D structure of camouflaged objects can be well predicted. Utilizing depth maps solely for auxiliary supervision will result in the omission of the powerful capacity of the depth estimator and the valuable information in the depth maps.

Therefore, in this paper, we aim to explore an effective and direct way of utilizing depth maps, while exploring a more reasonable multi-scale fusion approach to suppress the interference of inaccurate depth maps on camouflaged object detection. 
% Notably, we experimentally demonstrate that our method outperforms those presented in \cite{xiang2021exploring} even when using depth maps directly.
Notably, we experimentally demonstrate that our method outperforms those presented in \cite{xiang2021exploring, wu2023source} that also use depth prior.
On this basis, we propose a Depth-perceptual Attention Fusion Network (DAF-Net) for COD. 
Given that estimated depths in COD are often misleading, we specifically investigate the impact of low-quality depth maps on modal fusion. To this end, we propose a novel depth map constraint method with greater learning ability than previous fusion approaches. In the feature encoder stage, we extract and fuse RGB and depth information from each layer. Different backbones are employed for feature extraction to accommodate the distinct characteristics of each modality. For feature fusion, we propose a Depth-weighted Cross-attention Fusion (DCF) module that combines the depth and RGB maps. This module evaluates the importance of each modality during the fusion process, allowing for dynamic adjustment of the weights assigned to depth maps and RGB maps. Moreover, to capture comprehensive fusion information, we introduce Transformer as an auxiliary branch in this paper. The resulting three-branch encoder connects the RGB features, depth features and fusion features through the fusion module, facilitating mutual refinement among them. Additionally, we propose a simple Feature Aggregation decoder(FAD) that adaptively fuses the enhanced aggregated features while utilizing channel dependencies without increasing the model's complexity. Finally, we optimize the prediction map by using a cross-entropy loss on the output of each decoder.

Our main contributions are summarized as follows:

\begin{itemize}
\item The proposed DAF-Net has the characteristics of multi-scale and cross-modal feature weighted fusion, and employs depth information as auxiliary input. Even when only a limited area of depth map can highlight camouflaged objects, DAF-Net can still achieve more accurate segmentation of camouflaging object.
\item We investigate the effectiveness and rational use of depth information in the COD field. Our proposed Depth-weighted Cross-attention Fusion (DCF) module can effectively suppress redundant information and background noise while extracting valuable depth cues. Furthermore, our proposed Feature Aggregation Decoder (FAD) can adaptively fuse the enhanced aggregated features.
\item DAF-Net is evaluated on three widely used COD datasets, and the results demonstrate that it outperforms the 16 current state-of-the-art methods based on four commonly used evaluation metrics.
\end{itemize}

\section{Related work}
\subsection{RGB-D Salient Object Detection}

Substantial advancements have been achieved in existing SOD methods, but still fall short when dealing with challenging factors such as complex backgrounds or changing lighting conditions within a scene.
Recently, an increasing number of researchers have attempted to use depth maps as prior to supplement spatial information to the RGB image. Previous studies \cite{lang2012depth} have verified the significance of depth as a saliency cue in visual processing. Early RGB-D-based SOD models \cite{lang2012depth, ciptadi2013depth, peng2014rgbd} utilized manually extracted features, such as contrast and shape, to fuse depth and RGB information. However, traditional methods often lack of high-level reasoning required for complex scenes. 

Deep learning-based RGB-D SOD methods have demonstrated improved performance. Existing RGB-D SOD models can be categorized into three types in terms of the fusion strategies employed.
1) Early feature fusion, where RGB images and depth maps are directly concatenated and fed into the network\cite{song2017depth, liu2019salient, chen20223}.
2) Multi-scale fusion, where different levels of depth features are used to enhance RGB features by cross-modal weighted combination\cite{li2020icnet,ji2021calibrated,song2022improving}.
3) Late fusion, where RGB and depth maps are fed into two parallel networks respectively, and the learned features are concatenated at a later stage to generate the final saliency prediction \cite{han2017cnns, ding2019depth, 9585702, yao2022double, zhou2020attention}.
Among those works, multi-scale fusion exhibits better performance. 
Li et al. proposed ICNet \cite{li2020icnet}, while Ji et al.\cite{ji2021calibrated} suggested a depth calibration and fusion framework to enhance the performance of SOD with a learning strategy that calibrates the potential bias in the depth map. 
% Additionally, Song et al.\cite{song2022improving} proposed a modal-aware decoder that employs strategies such as feature embedding, modal inference, feature backprojection, and feature collection on the fused features.
Li et al. \cite{li2023robust}proposed a scribble-based weakly supervised RGB-D SOD method and devised corresponding solutions for the two inherent deficiencies of the suggested scribbles.
Additionally, Huang et al. \cite{huang2021employing} proposed a multi-modal feature interaction module and a saliency prior information guided fusion module to effectively capture cross-modal complementary information through a specialized fusion strategy.
In addition to conventional convolutional architectures, Liu et al. introduced a novel unified model based on a pure transformer (VST)\cite{liu2021visual}, which provides a new perspective for the SOD field.

\begin{figure*}[b]
\centerline{\includegraphics[width=1\textwidth]{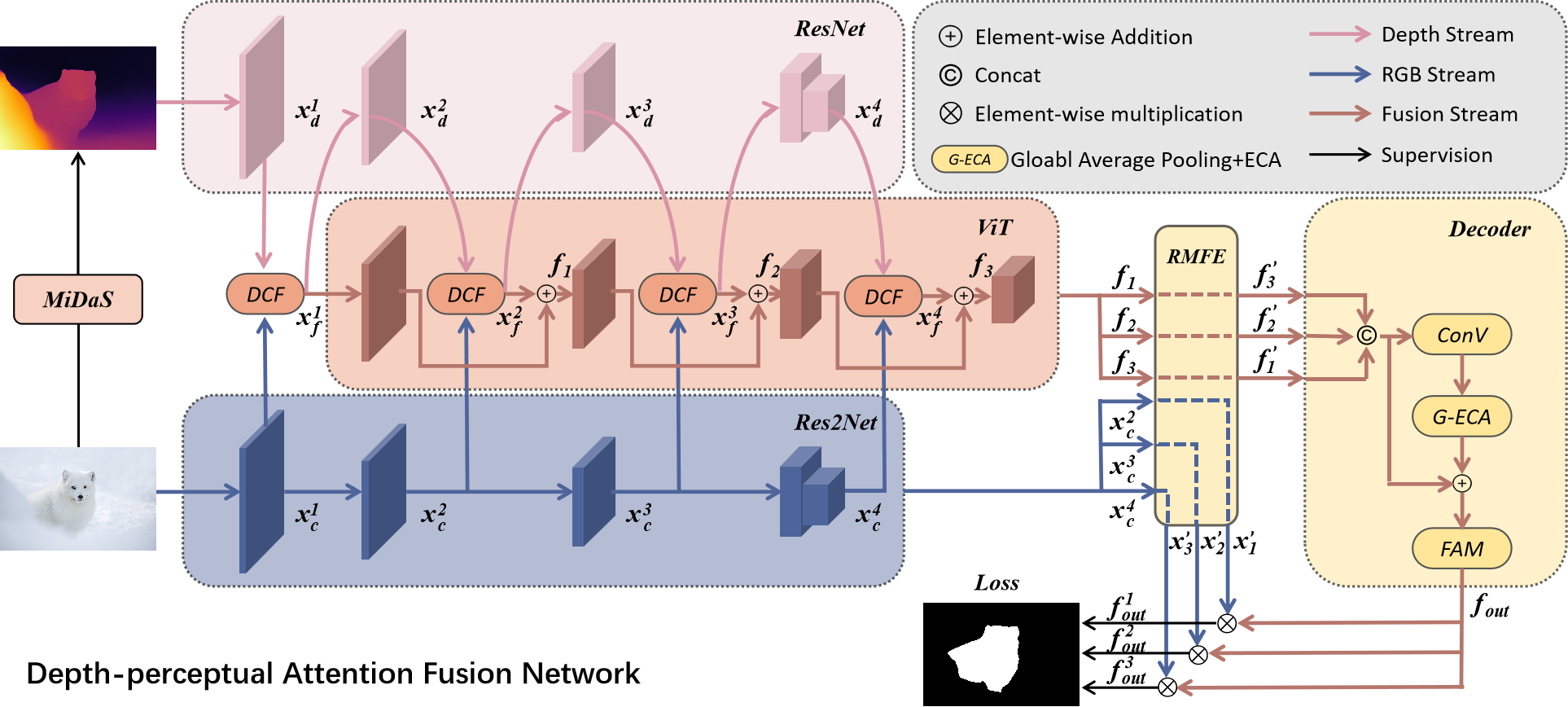}}
\caption{Overview of our proposed Depth-perceptual Attention Fusion Network(DAF-Net). The proposed network consists of our designed Depth-weighted Cross-attention Fusion (DCF, see Section \ref{AA}) module and Feature Aggregation Decoder (FAD, see Section \ref{BB}). DCF aims to fuse valuable depth cues while fully suppressing redundant information and background noise. The decoder aims to adaptively fuse the enhanced aggregated features without increasing the complexity of the model. RMFE is from \cite{zhu2022can}.}
\label{fig:fig2}
\end{figure*}

\subsection{RGB-D Camouflaged Object Detection}
Camouflaged object detection has a history dating back to 1998 when Tankus et al.\cite{tankus1998detection} proposed detecting artificial camouflaged objects in natural environments and combat scenarios using the non-edge region of interest mechanism. In recent years, deep learning-based methods for detecting camouflaged objects have demonstrated their strong feature extraction and autonomous learning abilities, resulting in continuous improvement in their detection accuracy. Some models\cite{mei2021camouflaged, fan2021concealed} adopt a coarse-to-fine strategy to initially predict the overall area and then gradually refine it, while others\cite{lv2021simultaneously, zhai2021mutual} use a multi-task learning strategy to extract richer information about camouflaged objects by collaborating across multiple tasks. However, existing models solely analyze cues from 2D images, disregarding the fact that visual systems function in 3D environment.

There is still a paucity of research in RGB-D COD, and we have conducted a thorough analysis of existing research
\cite{xiang2021exploring, wu2023source}.
DCNet\cite{xiang2021exploring} utilizes depth as an auxiliary branch rather than direct input. It is obvious that the auxiliary branch proposed in \cite{xiang2021exploring} has simpler network architecture and inferior capacity compared to sophisticated SIDE models such as MiDaS \cite{birkl2023midas}.
This simplicity might contribute to the relative mediocrity in the quality of the generated depth maps. This concern becomes particularly pronounced when there are no stringent evaluation criteria to rigorously control the depth map's quality during the fusion stage, potentially exerting a detrimental impact on the detection performance.
The recent PopNet \cite{wu2023source} predicts segmentation using a three-segment network of Depth Generation Network, Object Popping Network and Segmentation Network. Within the Object Popping Network, the depth map and RGB map are fused into the Popped-out Depth. Subsequently, the segmentation network utilizes the knowledge gained from the popped-out depth and RGB map to distinguish the object from the background.
In contrast to existing methods, our approach does not consider the depth maps as supervision for supplementary auxiliary branches, nor does it involve additional processing. Instead, we directly utilize the generated depth maps as input.
% Furthermore, considering the inherently high quality of the depth maps generated by the SIDE model, we advocate their direct utilization as inputs instead of employing them as supervision for supplementary auxiliary branches. 
This straightforward approach not only simplifies the model architecture but also harnesses the full potential of incorporated depth maps.
Considering the fact that there is no COD model that directly utilizes raw depth maps for detection tasks, we will explore an overall fusion segmentation network that utilizes the original depth map as input.

\subsection{Relationships Between RGB-D SOD and RGB-D COD}

Numerous previous works have demonstrated the effectiveness of depth in the SOD field, which prompts us to explore its potential in COD tasks. 
However, the accuracy of generated depth maps remains a primary concern. Even in the SOD work that uses sensored depth, some interference such as varying distances between the target and camera, occlusion by irrelevant objects and lighting conditions may result in noisy depth maps, and affect target detection. Effectively fusing RGB and depth images is crucial in RGB-D SOD models. Analogously, in the COD domain, minimizing the side effect of generated depth while maximizing its utility is of most importance.
% \textcolor{red}{
% However, SOD and COD tasks fundamentally differ in the context of the relationship between the target and its surrounding background. The heightened resemblance between a camouflage target and its environment renders the COD task more challenging than the SOD task. 
% Compounding this, the absence of depth datasets obtained from sensors in the COD field, coupled with the inevitable background interference in generated depth maps, underscores the increased complexity of RGB-D COD over RGB-D SOD.  
% These distinctions necessitate divergent emphases: in the RGB-D SOD model, the effective fusion of RGB and depth images holds paramount significance, while in the RGB-D COD domain, the primary focus is on mitigating the side effects of generated depth while maximizing its utility.
% }

\section{METHODOLOGY}

We propose to integrate generated depth cues as direct auxiliary input, thereby augmenting the network's capacity to detect camouflaged objects with 3D information.

\subsection{Depth Map Generation}
\begin{figure}[!t]
\centerline{\includegraphics[width=\columnwidth]{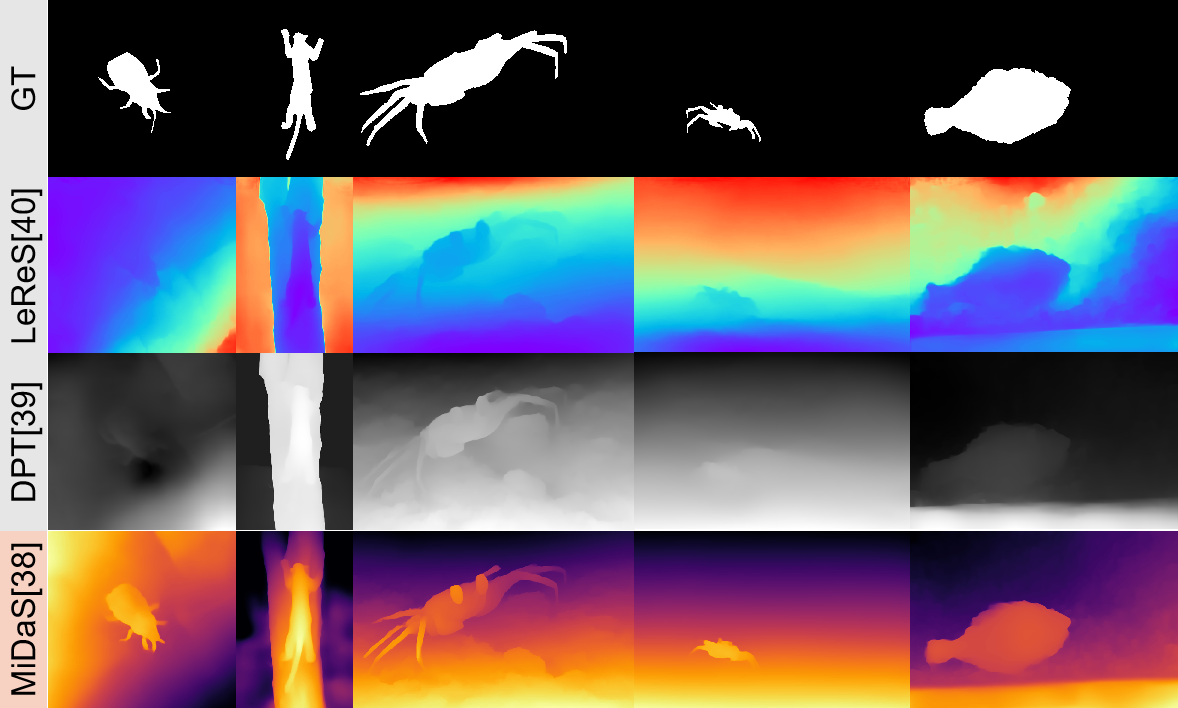}}
\caption{Visualization of depth maps generated by different advanced depth map estimation methods. Compared with DPT\cite{ranftl2021vision} and AdelaiDepth\cite{yin2022towards}, the depth maps produced by MiDaS\cite{birkl2023midas} can highlight the camouflaged objects best.}
\label{fig:fig3}
\end{figure}

Depth cues offer additional information about the visual content and can serve as a supplement to chromatic images in camouflaged object detection. In this study, we first investigate several state-of-the-art depth estimation methods, including DPT \cite{ranftl2021vision}, AdelaiDepth \cite{yin2022towards} and MiDaS v3.1 \cite{birkl2023midas} (which outperforms MiDaS v1.0 \cite{ranftl2020towards} used in \cite{xiang2021exploring}, and collectively referred to below as MiDaS). Figure \ref{fig:fig3} presents a qualitative comparison of these methods. Our findings agree with Xiang et al. \cite{xiang2021exploring} who chose MiDaS as their depth generator solely based on the visually plausible results.

Table \ref{tab:compare2} shows the results of our quantitative experiments. The comparison indicates that MiDaS \cite{birkl2023midas} produces better depth input to our COD model. We attribute the advantage of MiDaS to the utilization of its training set which comprises a vast array of natural scenes from movie footage. As described in \cite{birkl2023midas}, in addition to mixing existing datasets, the authors also collected a new 3D movie dataset which includes high-quality video frames of nature scenes with landscapes and animals in documentary features, resembling those images in COD datasets. This explains the reason why MiDaS outperforms the other depth estimators in RGB-D COD task. 
%effectively highlighting camouflages while minimizing noise. 
Therefore, we also select MiDaS as the depth estimator in our approach, and the generated depth map is directly fed into the proposed network as an auxiliary input.

It should be noted that generated depth suffers from limitations like inaccuracy and misalignment. We propose to address this issue by considering depth as supplementary information and seeking effective weights of depth features in conjunction with RGB features.

\subsection{Overall Architecture}
The structure of our proposed network model follows the encoder-decoder architecture paradigm, as shown in Figure \ref{fig:fig2}. The encoder extracts and fuses RGB and depth information, and the decoder aggregates these features and produces the prediction map.

In the encoder part, we employ CNN-based backbones as RGB feature extractor (main branch) and depth feature extractor (auxiliary branch). To capture a more comprehensive contextual understanding, we adopt a vision transformer architecture as the fusion subnetwork.
%We propose the Depth-weighted Cross-attention Fusion (DCF) module to fuse RGB and depth features, which controls the depth contribution and minimizes the interference from the noise of generated depth. As an additional enhancement, we integrated a ViT backbone into the fusion subnetwork, starting from the second layer of the encoder. By leveraging the broader receptive field of the Transformer, we were able to capture context features with enhancement by residual addition. 
Furthermore, in the fusion subnetwork, we design a Depth-weighted Cross-attention Fusion (DCF), which can control the depth contribution while fusing RGB and depth features to minimize the noise interference in the generated depth.
% \textcolor{red}{
% Note that in the RGB branch, the input of each layer of the network is derived from the output of the preceding layer within the same network. In contrast, in the depth branch, the input for each layer of the network originates from the output of the fusion module DCF.
% }

Three fused features from the final three layers of the fusion branch are chosen and refined by a feature enhancement module. 
These enhanced features are then input into the Feature Aggregation Decoder (FAD), where we employ a hybrid module with pooling and ECA \cite{wang2020eca} to explore the interdependencies among distinct channel features dynamically. Residual addition is used to further amplify the role of RGB features in generating the final prediction map. Detailed explanations of each component are in Sections \ref{AA}--\ref{CC}.

% Res2Net\cite{b34} and ResNet-50\cite{b35}
%In the decoder, we utilize feature maps generated by the last three layers of the fusion branch as input. Firstly, the feature enhancement module is employed to enhance these features. Subsequently, a combination module involving pooling and ECA \cite{b37} is used to adaptively explore the interdependence of different modalities. Lastly, the residual addition is applied to strengthen the contribution of RGB features in generating the final prediction map. Detailed explanations of each component are in Sections \ref{AA}--\ref{CC}.

\subsection{Feature Fusion module}\label{AA}

Figure \ref{fig:fig2} illustrates our proposed phased fusion strategy and Depth-weighted Cross-attention Fusion module. Unlike the commonly used average fusion which takes depth and RGB features of equal importance, we propose a depth-weighted fusion strategy that considers the quality of generated depth, aiming to effectively suppress noise and unrelated information while keeping valuable cues from the generated depth.
\begin{figure}[!t]
\centerline{\includegraphics[width=\columnwidth]{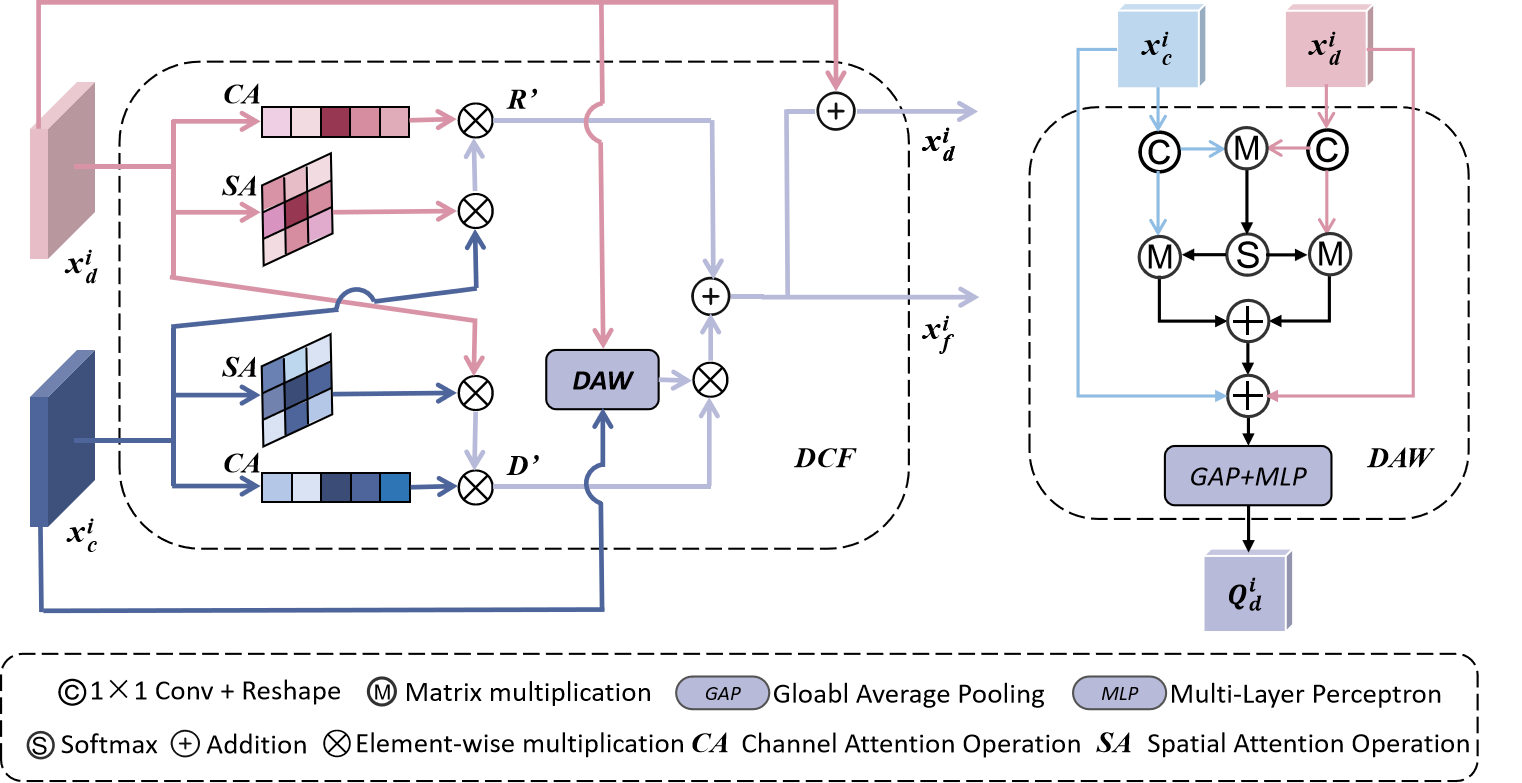}}
\caption{The details of Depth-weighted Cross-attention Fusion module (DCF).}
\label{fig:fig4}
\end{figure}

To achieve dynamic control over the contribution of depth features, we proposed Depth-weighted Cross-attention Fusion (DCF) module for accurately fusing the RGB and depth features from each backbone layer. 
We additionally introduce ``R50+ViT-B/16'' \cite{dosovitskiy2020image} with residual connections from the second layer of the auxiliary fusion branch to utilize the Transformer's wider receptive field to enhance context awareness.
The details of our DCF are depicted in Figure \ref{fig:fig4}. 
We utilize RGB features $x^{i}_{c} \in \mathbb{R}^{C_i \times H_i \times W_i}$ and depth features $x^{i}_{d} \in \mathbb{R}^{C_i \times H_i \times W_i}$ acquired from layer i ($i \in \{1, 2, 3, 4\}$) of the RGB backbone and depth backbone as input for DCF.
For each input feature map pair, we utilize channel attention to identify more important features and spatial attention to determine task-relevant regions \cite{woo2018cbam}. The cross-multiplication is then used to convert rich RGB/depth features into depth/RGB features, aiming to extract complementary information from the bimodal features. This process can be formulated as Eq. \ref{eq:fuseRD}.
\begin{equation}
\begin{gathered}
R'=(SA(x^{i}_{d}) \otimes x^{i}_{c}) \otimes CA(x^{i}_{d})\\
D'=(SA(x^{i}_{c}) \otimes x^{i}_{d}) \otimes CA(x^{i}_{c})
\label{eq:fuseRD}
\end{gathered}
\end{equation}
where 
% $x^{i}_{c}$ and $x^{i}_{d}$, $i\in\{1,2,3,4\}$ denote the input RGB and depth features from layer $i$, 
$SA(\cdot)$ and $CA(\cdot)$ denote the spatial attention and channel attention operations, $\otimes$ denotes the element-wise multiplication.

We employ the Deep Attention Weighting (DAW) module \cite{wu2022robust}, which provides an explicit weighting representation of the depth to control its influence in the model. 
The structure of the DAW module is depicted in Figure \ref{fig:fig4} (right). The input to DAW is also the RGB and depth features from the backbone encoders for each layer ($x^{i}_{c} \in \mathbb{R}^{C_i \times H_i \times W_i}$ and $x^{i}_{d} \in \mathbb{R}^{C_i \times H_i \times W_i}$). The output of DAW is a confidence value, $Q^{i}_{d}$, which serves as a weight to adjust the contribution of depth features. The entire process can be formulated as Eq.\ref{eq:daw1}, Eq.\ref{eq:daw2} and Eq.\ref{eq:daw3}:
% \begin{equation}
\begin{align}
\label{eq:daw1}
&F_a=\sigma(RS_d(Conv(x^{i}_{d})) \times RS_c(Conv(x^{i}_{c})))\\
\label{eq:daw2}
&F_{a2}=F_a \times Conv(x^{i}_{d})+F_a \times Conv(x^{i}_{c})\\
\label{eq:daw3}
&Q^{i}_d=MLP(GAP(F_{a2}+(x^{i}_{d}+x^{i}_{c})))
\end{align}
% \end{equation}
where $\sigma(\cdot)$ denotes the softmax operation, $RS_d$ and $RS_c$ denote the reshape operations on the depth and RGB maps, the purpose is to turn them into the size of $C \times HW$, $Conv(\cdot)$ denotes a 1x1 convolutional layer, 
$GAP(\cdot)$ denotes global average pooling operation and $MLP(\cdot)$ denotes multilayer perception. 
Finally, we apply the acquired weights to the attention-processed depth features obtained through Eq.\ref{eq:fuseRD}, yielding the corresponding fusion map $x^{i}_f$.  
We sum the fusion map $x^{i}_f$ and depth map $x^{i}_{d}$ as the input to the i+1 layer of the depth subnetwork.
The entire process can be formulated as Eq.\ref{eq:final1} and Eq.\ref{eq:final2}.
% \begin{equation}
\begin{align}
\label{eq:final1}
&x^{i}_f=Q^{i}_d\otimes D'+ R’\\
\label{eq:final2}
&x^{i}_d=x^{i}_{d} + x^{i}_f
\end{align}
% \end{equation}
% The final fusion of depth and RGB features is $x^{i}_f$, which considers depth contribution.

\subsection{Feature Aggregation Decoder} \label{BB}

The global information is more important in COD as the model needs to distinguish the camouflaged objects from their surroundings. Before feeding to the decoder, the fused features are first enhanced by the Residual Multi-scale Feature Extractor (RMFE) module \cite{zhu2022can}.
Those enhanced features are then passed into our proposed Feature Aggregation Decoder, which is illustrated in Figure \ref{fig:fig2}. The decoder comprises a vanilla Efficient Channel Attention (ECA) module \cite{wang2020eca} and a Feature Aggregation Module (FAM) \cite{liu2019simple}.

Concretely, the three fused features from the last three stages of the fusion subnetwork ($f_1$, $f_2$ and $f_3$) are first concatenated and then sent to a combination module of pooling and ECA to dynamically explore the interdependence among distinct channel features. The FAM is employed to mitigate the aliasing effect and enhance the detection performance, after which the residual addition is incorporated. The entire process can be formulated as Eq.\ref{eq:rmfe1}, Eq.\ref{eq:decoder1} and Eq.\ref{eq:decoder2}:
\begin{align}
\label{eq:rmfe1}
&f_3', f_2', f_1'=RMFE(f_3, f_2, f_1)\\
\label{eq:decoder1}
&F=cat(f_3', f_2', f_1')\\
\label{eq:decoder2}
&f_{out}=FAM(GECA(F)+F)
\end{align}
% where the inputs $f_3$, $f_2$ and $f_1$ are three fused features obtained from different stages, 
where $RMFE(\cdot)$ denotes the RMFE module used to perform feature augmentation, $cat(\cdot)$ denotes the concatenation operation, $GECA(\cdot)$ represents vanilla ECA module and $FAM(\cdot)$ represents feature aggregation module. 

We further improve the weight of RGB features by element-wise multiplying the decoded fusion features with the enhanced RGB features.The entire process can be formulated as Eq.\ref{eq:rmfe2} and Eq.\ref{eq:res}:
\begin{align}
\label{eq:rmfe2}
&x_3',x_2',x_1'=RMFE(x^4_c,x^3_c,x^2_c) \\
% \begin{equation}
&
\begin{gathered}
% \label{eq:res1}
f^1_{out}=Conv_1(f_{out}) \otimes x_1' \\
% \label{eq:res2}
f^2_{out}=Conv_2(f_{out}) \otimes x_2' \\
f^3_{out}=Conv_3(f_{out}) \otimes x_3' 
\label{eq:res}
\end{gathered}
\end{align}
% \end{equation}
where $Conv_1(\cdot)$, $Conv_2(\cdot)$, $Conv_3(\cdot)$ stand for convolutions 
% to size different layer features.
to align the spatial sizes of the different layer features.
The three outputs $f^1_{out}$, $f^2_{out}$ and $f^3_{out}$ are supervised by the ground truth after an upsampling operation, where $f^1_{out}$ is used to generate the final prediction.

%The calibrated and augmented $F^1_{out}$ is subsequently input into the classifier layer to generate the final prediction.

\subsection{Loss function} \label{CC}
%The decoder's three output features for camouflaged object prediction are supervised by the ground truth after an upsampling operation. 
Following\cite{fan2021concealed}, we use binary cross-entropy loss ($L_{BCE}$) for pixel-level restriction and intersection over union loss ($L_{IoU}$) for image-level restriction in each layer output. The loss output is represented as Eq.\ref{eq:loss1}:
\begin{equation}
L= L_{BCE} + L_{IoU}
\label{eq:loss1}
\end{equation}

Finally, the total training loss for the proposed model can be represented as Eq.\ref{eq:lossTotal}:
\begin{equation}
L_{total}=\sum_{i=1}^{3} \lambda _i L(f^i_{out},G)
\label{eq:lossTotal}
\end{equation}
where $f^i_{out}$, $i \in \{1,2,3\}$ and $G$ represent the $i$-th predicted camouflaged map and ground truth respectively, $f^1_{out}$ denotes the final camouflaged prediction, $\lambda _i$ is a weight set according to the spatial sizes of the final output features, in this paper we set as \{0.25, 0.5, 1\}.

\begin{table*}[!t]
\caption{Quantitative comparison with 16 COD methods and 7 SOD methods on three datasets. The best results are highlighted in \textbf{Bold}.}
\resizebox{\textwidth}{!}{
\begin{tabular}{|c|c|cccc|cccc|cccc|}
\hline
\multirow{2}{*}{\centering \textbf{Model}}&\multirow{2}{*}{\centering \textbf{Pub/Year}}&\multicolumn{4}{|c|}{\textbf{CAMO-Test}}&\multicolumn{4}{|c|}{\textbf{COD10K-Test}}&\multicolumn{4}{|c|}{\textbf{NC4K}} \\
\cline{3-14} 
&&\textbf{\textit{$S_\alpha\uparrow$}}& \textbf{\textit{$E_\phi\uparrow$}}& \textbf{\textit{$F^\omega _\beta\uparrow$}}& \textbf{\textit{$\mathcal{M}\downarrow$}}&\textbf{\textit{$S_\alpha\uparrow$}}& \textbf{\textit{$E_\phi\uparrow$}}& \textbf{\textit{$F^\omega _\beta\uparrow$}}& \textbf{\textit{$\mathcal{M}\downarrow$}}&\textbf{\textit{$S_\alpha\uparrow$}}& \textbf{\textit{$E_\phi\uparrow$}}& \textbf{\textit{$F^\omega _\beta\uparrow$}}& \textbf{\textit{$\mathcal{M}\downarrow$}} \\
\hline
\multicolumn{14}{|c|}{\centering \textbf{RGB-D SOD Models}}\\
\hline
S$^2$MA\cite{9585702}&TPAMI$_{21}$&0.782&0.834&0.675&0.079&0.735&0.796&0.539&0.055&0.772&0.823&0.645&0.073 \\
DFM-Net\cite{zhang2021depth}&MM$_{21}$ &0.799&0.818&0.664&0.079&0.751&0.772&0.508&0.055&0.801&0.820&0.646&0.069 \\
SP-Net\cite{zhou2021specificity}& ICCV$_{21}$   &0.831&0.877&0.767&0.060&0.825&0.888&0.705&0.032&0.844&0.893&0.771&0.047 \\
VST\cite{liu2021visual}&ICCV$_{21}$&0.829&0.863&0.739&0.060&0.803&0.841&0.624&0.039&0.848&0.881&0.744&0.046 \\
CIRNet\cite{cong2022cir}&TIP$_{22}$&0.797&0.802&0.681&0.080&0.813&0.832&0.638&0.038&0.835&0.852&0.723&0.055 \\
MaDNet\cite{song2022improving}&TIP$_{22}$&0.742&0.767&0.583&0.111&0.780&0.763&0.471&0.067&0.761&0.793&0.588&0.087 \\
HINet\cite{bi2023cross}& PR$_{22}$      &0.820&0.846&0.716&0.069&0.785&0.813&0.588&0.044&0.818&0.844&0.693&0.061 \\
\hline
\multicolumn{14}{|c|}{\centering \textbf{COD Models}}\\
\hline
SINet\cite{fan2020camouflaged}& CVPR$_{20}$&0.751&0.771&0.606&0.100&0.771&0.806&0.551&0.051&0.808&0.871&0.723&0.058 \\
TANet\cite{ren2021deep}&
TCSVT$_{21}$&0.793&0.834&0.690&0.083&0.803&0.848&0.629&0.041&-&-&-&-\\
PFNet\cite{mei2021camouflaged}& CVPR$_{21}$&0.782&0.841&0.695&0.085&0.800&0.877&0.660&0.040&0.829&0.887&0.745&0.053\\
UGTR\cite{yang2021uncertainty}& ICCV$_{21}$&0.784&0.822&0.684&0.086&0.817&0.853&0.666&0.036&0.839&0.875&0.747&0.052\\
DCNet\cite{xiang2021exploring}&arXiv$_{22}$&0.819&0.881&–&0.069&0.829&\textbf{0.903}&–&0.032&0.855&0.910&–&\textbf{0.042}\\
PreyNet\cite{zhang2022preynet}&MM$_{22}$&0.813&0.876&0.748&0.071&0.830&0.895&0.719&0.032&–&–&–&–\\
BSANet\cite{zhu2022can}&AAAI$_{22}$ &0.794&0.851&0.717&0.079&0.818&0.891&0.699&0.034&0.841&0.897&0.771&0.048\\
OCENet\cite{liu2022modeling}&WACV$_{22}$&0.802&0.852&0.723&0.080&0.827&0.894&0.707&0.033&0.853&0.902&0.785&0.045\\
BGNet\cite{sun2022boundary}&IJCAI$_{22}$&0.812&0.870&0.749&0.073&0.831&0.901&0.722&0.033&0.851&0.907&0.788&0.044\\
SegMaR\cite{jia2022segment}&CVPR$_{22}$&0.815&0.874&0.753&0.071&0.833&0.899&\textbf{0.724}&0.034&0.841&0.896&0.781&0.046\\
SINetV2\cite{fan2021concealed}&TPAMI$_{22}$&0.820&0.882&0.743&0.070&0.815&0.887&0.680&0.037&0.847&0.903&0.770&0.048\\
C$^2$F-Net\cite{chen2022camouflaged}&TCSVT$_{22}$&0.800&0.869&0.730&0.077&0.811&0.891&0.691&0.036&-&-&-&-\\
LSR+\cite{lv2023towards}&TCSVT$_{23}$&0.820&0.874&–	&0.071&0.831&0.898&	–	&0.032&0.859&\textbf{0.911}&–	&\textbf{0.042}\\
PFNet+\cite{mei2023distraction}&SCIS$_{23}$&0.791&0.850&0.713&0.080&0.806&0.884&0.677&0.037&–&–&–&–\\
DGNet\cite{ji2023deep}&MIR$_{23}$ &0.839&0.901&0.769&0.057&0.822&0.896&0.693&0.033&0.857&\textbf{0.911}&0.784&\textbf{0.042}\\
PopNet\cite{wu2023source}&ICCV$_{23}$ &0.806&0.869&-&0.073&0.827&0.897&-&\textbf{0.031}&0.852&0.908&-&0.043\\
\hline
DAF-Net		     &&\textbf{0.860}&\textbf{0.913}&\textbf{0.799}&\textbf{0.051}&\textbf{0.838}&0.899&0.715&\textbf{0.031}&\textbf{0.865}&0.909&\textbf{0.792}&\textbf{0.042}\\
\hline
\end{tabular}}
\label{tab:compare1}
\end{table*}
\begin{table*}[!t]
\caption{Quantitative comparison of depth maps generated by different depth estimation models. The best results are highlighted in \textbf{Bold}.}
\resizebox{\textwidth}{!}{
\begin{tabular}{|c|c|cccc|cccc|cccc|}
\hline
\multirow{2}{*}{\centering \textbf{Model}}&\multirow{2}{*}{\centering \textbf{Pub/Year}}&\multicolumn{4}{|c|}{\textbf{CAMO-Test}}&\multicolumn{4}{|c|}{\textbf{COD10K-Test}}&\multicolumn{4}{|c|}{\textbf{NC4K}} \\
\cline{3-14} 
&&\textbf{\textit{$S_\alpha\uparrow$}}& \textbf{\textit{$E_\phi\uparrow$}}& \textbf{\textit{$F^\omega _\beta\uparrow$}}& \textbf{\textit{$\mathcal{M}\downarrow$}}&\textbf{\textit{$S_\alpha\uparrow$}}& \textbf{\textit{$E_\phi\uparrow$}}& \textbf{\textit{$F^\omega _\beta\uparrow$}}& \textbf{\textit{$\mathcal{M}\downarrow$}}&\textbf{\textit{$S_\alpha\uparrow$}}& \textbf{\textit{$E_\phi\uparrow$}}& \textbf{\textit{$F^\omega _\beta\uparrow$}}& \textbf{\textit{$\mathcal{M}\downarrow$}} \\
\hline
\textbf{MiDaS \cite{birkl2023midas} + DAF-Net}& arXiv$_{23}$ &\textbf{0.860}&\textbf{0.913}&\textbf{0.799}&\textbf{0.051}&\textbf{0.838}&\textbf{0.899}&\textbf{0.715}&\textbf{0.031}&\textbf{0.865}&\textbf{0.909}&\textbf{0.792}&\textbf{0.042}\\
DPT\cite{ranftl2021vision} + DAF-Net&ICCV$_{21}$ &0.817&0.864&0.735&0.070&0.834&0.890&0.709&0.033&0.861&0.905&0.786&0.043 \\
LeReS\cite{yin2022towards} + DAF-Net& TPAMI$_{22}$&0.830&0.875&0.754&0.066&0.835&0.897&0.708&0.032&0.861&0.905&0.785&0.043 \\
\hline
\end{tabular}}
\label{tab:compare2}
\end{table*}

\section{EXPERIMENTS}
In this section, we begin by providing an overview of the experiments, which includes details about the training/testing dataset and the validation metrics used. Following that, we perform quantitative and qualitative comparison experiments involving 16 well-established state-of-the-art COD methods and 7 RGB-D based SOD methods. Lastly, we conduct ablation experiments on each of our proposed modules to validate the effectiveness of the components integrated into the proposed framework.

\subsection{Datasets and evaluation metrics}
\subsubsection{Datasets}
We evaluate our method on three benchmark datasets: CAMO\cite{le2019anabranch}, COD10K\cite{fan2020camouflaged} and NC4K\cite{lv2021simultaneously}. CAMO is widely recognized and extensively utilized for camouflaged object detection. It comprises 1000 training images and 250 test images, encompassing both natural camouflage (camouflaged animals) and artificial camouflage (body painting and military camouflage). Recognizing these camouflaged categories presents significant difficulties. COD10K stands out as the most challenging camouflaged object detection dataset to date, with 3040 training images and 2026 test images. This dataset has significantly contributed to the advancement of camouflaged object detection techniques. NC4K currently holds the distinction of being the largest camouflaged object test set and is widely employed for model evaluation. It consists of 4121 images downloaded from the Internet, predominantly featuring natural camouflaged categories, although a small proportion includes instances of artificial camouflage.

\begin{figure*}[!t]
\centerline{\includegraphics[width=1\textwidth]{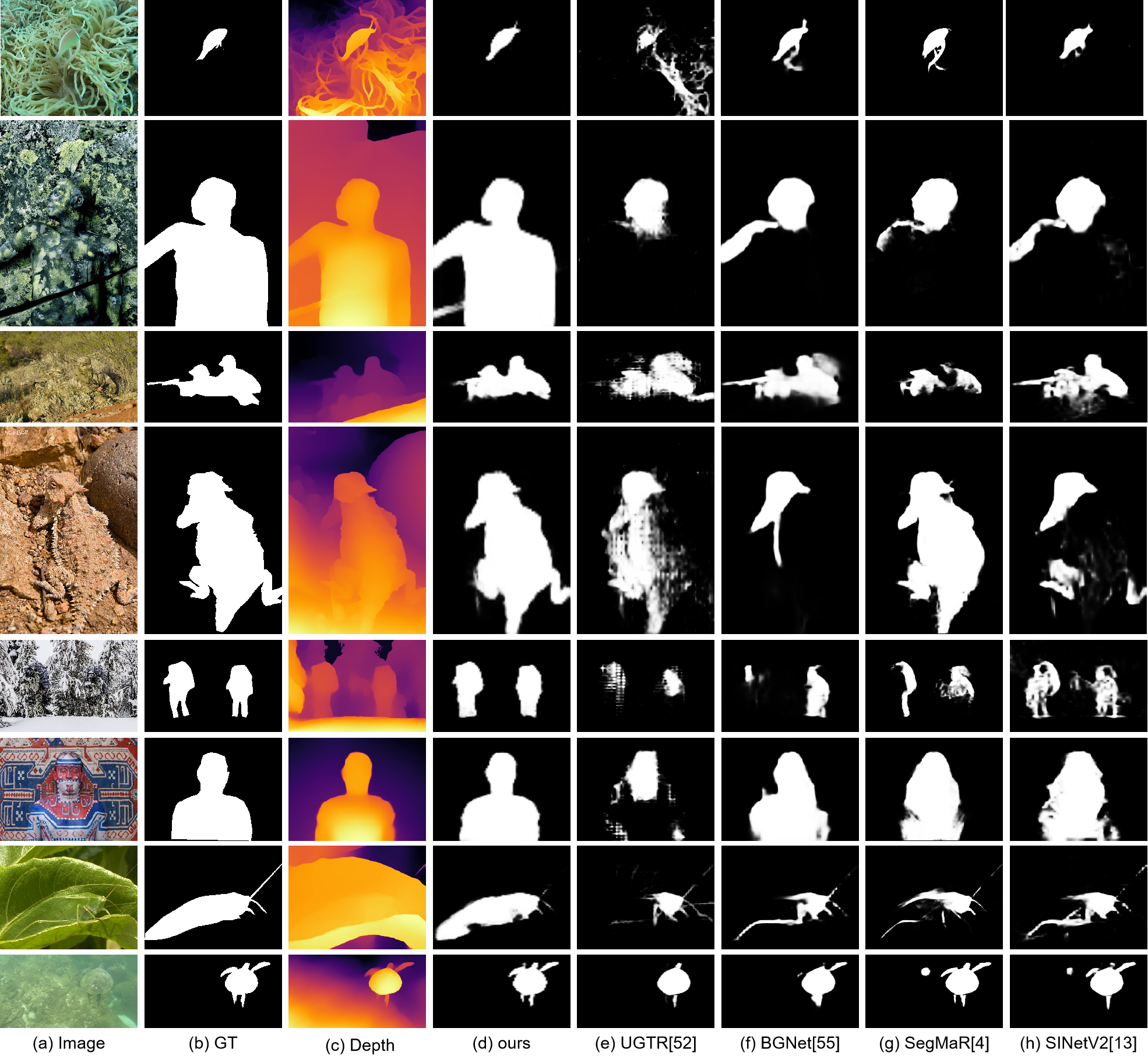}}
\caption{Qualitative comparison with eight state-of-art COD methods.}
\label{fig:results}
\end{figure*}

To ensure consistency with previous studies, we utilized 3040 samples from COD10K and 1000 samples from CAMO as the training set, while the test set consists of 2026 test images under COD10K, 250 test images under CAMO and the entire NC4K dataset.

\subsubsection{Evaluation metrics}
We use four metrics that are commonly used in COD tasks to evaluate the model performance: S-measure ($S_\alpha$), weighted F-measure ($F^\omega _\beta$), E-measure ($E_\phi$) and Mean absolute error ($\mathcal{M}$). We utilize $F^\omega_\beta$ to determine the weighted relationship between precision and recall. $S_\alpha$ is employed to assess the structural similarity between the predicted map and the ground truth. Furthermore, $E_\phi$ is employed to evaluate the global and local accuracy of the camouflaged object detection results by comparing the discrepancies between the predicted and true maps. Lastly, $\mathcal{M}$ is utilized to compute the mean absolute error for each pixel.

\subsection{Implementation Details}
The proposed DAF-Net is implemented with PyTorch. The DAF-Net uses Res2Net\cite{gao2019res2net}, ResNet-50\cite{he2016deep} and R50+ViT-B/16\cite{dosovitskiy2020image} pretrained on ImageNet as backbone. AdamW with an initial learning rate of $1e-5$ is chosen as the optimizer. The learning rate is divided by 10 after every 50 epochs. During the training phase, we utilize a dataset comprising 4040 camouflaging object images extracted from CAMO and COD10K. To ensure uniformity, all input images are resized to 352$\times$352. Furthermore, in order to mitigate overfitting, we apply multiple augmentation strategies, including random horizontal flipping and boundary cropping, to the input images. During the testing phase, we resize the input images from CAMO, NC4K and COD10K to the same dimensions of $352 \times 352$, and are then restored to their original sizes for evaluation purposes. The entire model is trained for 100 epochs with a batch size of 4 on a single NVIDIA 3090 GPU.

\begin{table*}[!t]
\caption{Ablation study of the proposed network. DCF: Depth-weighted Cross-attention Fusion module. RMFE: Residual Multi-scale Feature Extractor module. FAD: Feature Aggregation Decoder. The best result is highlighted in \textbf{Bold}}.
\centering
\resizebox{\textwidth}{!}{
\begin{tabular}{c|cccc|cccc|cccc}
\hline
\multirow{2}{*}{No} & \multicolumn{4}{c|}{Component} & \multicolumn{4}{c}{CAMO-Test} & \multicolumn{4}{c}{COD10K-Test} \\ \cline{2-13} 
&Encoder architecture&DCF   &RMFE  & FAD  &$S_\alpha\uparrow$&$F_\beta^w\uparrow$&$E_\phi\uparrow$&$\mathcal{M}\downarrow$&$S_\alpha\uparrow$&$F_\beta^w\uparrow$&$E_\phi\uparrow$&$\mathcal{M}\downarrow$\\ \hline
\#1 & ResNet50+ViT+ResNet50&  &   &       & 0.831 & 0.753 & 0.891 & 0.059 & 0.803 & 0.655 & 0.873 & 0.038\\ 
\#2 & ResNet50+ViT+ResNet50&  &   &\checkmark& 0.854 & 0.788 & 0.903 & 0.052 & 0.822 & 0.688 & 0.884 & 0.034\\ 
\#3 & ResNet50+ViT+ResNet50&\checkmark& &    & 0.851 & 0.781 & 0.895 & 0.056 & 0.827 & 0.693 & 0.889 & 0.033 \\ 
% \#4 & ResNet50+ViT+ResNet50&\checkmark& &\checkmark& 0.859 & 0.796 & 0.908 & 0.051 & 0.825 & 0.693 & 0.888 & 0.033\\
\#4 & ResNet50+ViT+ResNet50&\checkmark& &\checkmark& 0.853 & 0.783 & 0.901 & 0.054 & 0.825 & 0.693 & 0.888 & 0.033\\
\#5 & ResNet50+ViT+ResNet50&\checkmark& \checkmark&\checkmark& 0.856 & 0.789 & 0.903 & \textbf{0.051} & 0.828 & 0.698 & 0.890 & 0.032 \\
\#6 & Res2Net+ViT+Res2Net  &\checkmark& \checkmark&\checkmark& 0.859 & 0.796 & 0.909 & 0.052 & \textbf{0.838} & 0.713 & \textbf{0.899} & \textbf{0.031} \\ 
\#7 & ResNet50+ViT+Res2Net &\checkmark& \checkmark&\checkmark& \textbf{0.860} & \textbf{0.799} & \textbf{0.913} & \textbf{0.051} & \textbf{0.838}& \textbf{0.715} & \textbf{0.899} & \textbf{0.031} \\ 
\hline
\end{tabular}
}
\label{tab:ablationall}
\end{table*}
% \begin{table}[!t]
% \caption{Ablation studies on two datasets. The best results are highlighted in \textbf{Bold}.}
% \centering
% \resizebox{\columnwidth}{!}{
% \begin{tabular}{c|cccc|cccc}
% \hline
% \multirow{2}{*}{No.} &\multicolumn{4}{c|}{CAMO-Test} &\multicolumn{4}{c}{COD10K-Test}  \\ 
% \cline{2-9} \multicolumn{1}{c|}{} &$S_\alpha\uparrow$ & $F_\beta^w\uparrow$ & $E_\phi\uparrow$  & $\mathcal{M}\downarrow$ & $S_\alpha\uparrow$ &$F_\beta^w\uparrow$ & $E_\phi\uparrow$ & $\mathcal{M}\downarrow$  \\ \hline
% \#1 & 0.831 & 0.753 & 0.891 & 0.059 & 0.803 & 0.655 & 0.873 & 0.038 \\ 
% \#2 & 0.854 & 0.788 & 0.903 & 0.052 & 0.822 & 0.688 & 0.884 & 0.034 \\ 
% \#3 & 0.851 & 0.781 & 0.895 & 0.056 & 0.827 & 0.693 & 0.889 & 0.033 \\ 
% \#4 & 0.859 & 0.796 & 0.908 & 0.051 & 0.825 & 0.693 & 0.888 & 0.033 \\ 
% \#5 & 0.856 & 0.789 & 0.903 & 0.051 & 0.828 & 0.698 & 0.890 & 0.032 \\ 
% \#6 & 0.859 & 0.796 & 0.909 & 0.052 & \textbf{0.838} & 0.713 & \textbf{0.899} & \textbf{0.031} \\ 
% \#7 & \textbf{0.860} & \textbf{0.799} & \textbf{0.913} & \textbf{0.051} & \textbf{0.838}& \textbf{0.715} & \textbf{0.899} & \textbf{0.031} \\ 
% \hline
% \end{tabular}}
% \label{tab:ablation2}
% \end{table}

\subsection{Comparison with state-of-the-arts}
To demonstrate the effectiveness of depth maps and the superiority of our proposed method, we conducted a comparison with 16 mainstream methods, which included 14 common camouflaged object detection methods (SINet \cite{fan2020camouflaged}, TANet\cite{ren2021deep} PFNet \cite{mei2021camouflaged}, UGTR \cite{yang2021uncertainty}, PreyNet \cite{zhang2022preynet}, BSANet \cite{zhu2022can}, OCENet \cite{liu2022modeling}, BGNet \cite{sun2022boundary}, SegMaR \cite{jia2022segment}, SINetV2 \cite{fan2021concealed}, C$^2$F-Net\cite{chen2022camouflaged}, LSR+ \cite{lv2023towards}, PFNet+ \cite{mei2023distraction}, DGNet \cite{ji2023deep}) and two camouflaged object detection methods incorporating depth cues (DCNet \cite{xiang2021exploring}, PopNet \cite{wu2023source}). All predictions made by the competitors were either provided by the authors or generated by models retrained on open-source code.

Due to the scarcity of two-branch models trained with depth maps in the domain of camouflaged object detection, we choose 7 RGB-D models from the field of salient object detection (S$^2$MA\cite{9585702}, DFM-Net\cite{zhang2021depth}, SP-Net\cite{zhou2021specificity}, VST\cite{liu2021visual}, CIRNet\cite{cong2022cir}, HINet\cite{bi2023cross}, MaDNet\cite{song2022improving}) that possess similar structures. These selected models are then trained and fine-tuned on the COD training set to facilitate a comprehensive comparative analysis. To ensure a fair evaluation, all models undergo 100 epochs of training, while maintaining the original settings of each model.

\subsubsection{Quantitative evaluation}
Table \ref{tab:compare1} presents the quantitative results of the proposed method alongside 16 state-of-the-art methods for camouflaged object detection and 7 methods for salient object detection on three benchmark datasets. The results indicate that our proposed method outperforms other models across most evaluation indicators for all datasets. Notably, on the CAMO dataset, our proposed method demonstrates a notable improvement in $F^\beta$ and $S_\alpha$ metrics, with an increase of 3.0$\%$ and 2.1$\%$, respectively, compared to the suboptimal method DGNet \cite{ji2023deep}. Moreover, when compared to DCNet \cite{xiang2021exploring} and PopNet\cite{wu2023source}, which also incorporate deep prior, our proposed method achieves a significant improvement of 4.1$\%$, 0.9$\%$, 1.0$\%$ and 6.0$\%$, 1.1$\%$, 1.3$\%$ in $S_\alpha$ across the respective datasets.

Furthermore, we conduct a comparison of the depth map quality generated by three different single image depth estimation (SIDE) models (DPT \cite{ranftl2021vision}, AdelaiDepth \cite{yin2022towards} and MiDaS \cite{birkl2023midas}). 
Due to the absence of precise depth maps obtained from specialized cameras for camouflaged object detection, we employ depth maps generated by various SIDE models as inputs for our DAF network. The evaluation of these depth maps is centered on their complementary impact on the overall model performance. The quantitative results of the three selected advanced depth estimation models on the three benchmark datasets are presented in Table \ref{tab:compare2}. Notably, the depth map generated by the MiDaS, which was chosen for this study, outperformed other models across all four evaluation indicators in the three datasets.

\subsubsection{Qualitative comparisons}
Figure \ref{fig:results} illustrates challenging scenarios and presents the results obtained from our method and other top models. We conduct a visual comparison of eight test samples obtained from the COD10K-Test and NC4K datasets, including UGTR, BgNet, SegMaR, SINetV2 and our proposed DAF-Net. These samples represent a range of challenging scenarios, including targets with complex structures and boundaries (rows 3 and 4), multiple targets (row 5), targets seamlessly integrated into the surrounding environment (rows 2, 6 and 7) and targets in complex environments such as underwater (rows 1 and 8). In contrast to most existing methods that struggle to detect the completely camouflaged object, our method excels in accurately discovering the entire target concealed within the environment. This remarkable performance is attributed to the effective utilization of depth maps.

Moreover, Figure \ref{fig:results} shows that while some depth maps are not as effective at highlighting the camouflaged targets contained within them, they do contain valuable contour information. This continuous contour information is instrumental in facilitating the model's accurate detection of the complete target.

\subsection{Ablation study}
Ablation studies have been conducted to verify the effectiveness of each major component in the proposed DAF-Net. The quantitative results of these studies are presented in Table \ref{tab:ablationall}, Table \ref{tab:ablation2}, Table \ref{tab:ablation3} and Table \ref{tab:ablation5}.

\noindent{\textbf{Effectiveness of the DCF.}}
To extract valuable depth cues while effectively suppressing redundant information, we propose a depth-weighted cross-attention fusion module. We present experimental results from Table \ref{tab:ablationall} to demonstrate the effectiveness of this module.
Moreover, a series of experiments are conducted to validate the efficacy of the different components and designs incorporated in the DCF module. The outcomes of these experiments are presented in Table \ref{tab:ablation2}.

We start by removing all critical modules, using a three-branch architecture and base fusion modules for fusion, and adopting simple concat and convolution as decoder. As shown in Table \ref{tab:ablationall} the baseline version (denoted as \#1) yields the lowest scores across all evaluation metrics. 
Building upon this baseline, to assess the effectiveness of the DCF module in our model, we conduct two sets of ablation experiments: \#1 and \#3, \#2 and \#4 as shown in Table \ref{tab:ablationall}. The results from \#1 and \#3 demonstrate that the inclusion of DCF enhances the model's performance across all four metrics compared to the baseline. Additionally, the results from \#2 and \#4 indicate that combining DCF with the Feature Aggregation Decoder enhances the model's capability to detect camouflaging objects.

% \begin{table*}[!t]
% \caption{Ablation study of the strategy of using different depth maps to generate depth map weights in the DCF module on three datasets. The best result is highlighted in \textbf{bold}.}
% \resizebox{\textwidth}{!}{
% \begin{tabular}{|c|cccc|cccc|cccc|}
% \hline
% \multirow{2}{*}{\centering \textbf{}}&\multicolumn{4}{|c|}{\textbf{CAMO-Test}}&\multicolumn{4}{|c|}{\textbf{COD10K-Test}}&\multicolumn{4}{|c|}{\textbf{NC4K}} \\
% \cline{2-13} 
% &\textbf{\textit{$S_\alpha\uparrow$}}& \textbf{\textit{$E_\phi\uparrow$}}& \textbf{\textit{$F^\omega _\beta\uparrow$}}& \textbf{\textit{$\mathcal{M}\downarrow$}}&\textbf{\textit{$S_\alpha\uparrow$}}& \textbf{\textit{$E_\phi\uparrow$}}& \textbf{\textit{$F^\omega _\beta\uparrow$}}& \textbf{\textit{$\mathcal{M}\downarrow$}}&\textbf{\textit{$S_\alpha\uparrow$}}& \textbf{\textit{$E_\phi\uparrow$}}& \textbf{\textit{$F^\omega _\beta\uparrow$}}& \textbf{\textit{$\mathcal{M}\downarrow$}} \\
% \hline
% First Layer  &0.849&0.897&0.779&0.057&0.839&0.898&0.712&0.031&0.866&0.910&0.790&0.041 \\
% Second Layer &0.856&0.901&0.783&0.056&0.837&0.898&0.709&0.031&0.863&0.906&0.784&0.043\\
% Third Layer  &0.854&0.904&0.789&0.053&0.836&0.898&0.710&0.031&0.863&0.907&0.787&0.042 \\
% ALL Layer	 &0.860&0.913&0.799&0.051&0.838&0.899&0.715&0.031&0.865&0.909&0.792&0.042\\
% \hline
% \end{tabular}}
% \label{tab:ablation4}
% \end{table*}

\begin{table*}[!t]
\caption{Ablation study of Depth-weighted Cross-attention Fusion module on three datasets. The best result is highlighted in \textbf{Bold}.}
\resizebox{\textwidth}{!}{
\begin{tabular}{cc|cccc|cccc|cccc|}
\hline
\multicolumn{2}{|c|}{\multirow{2}{*}{\textbf{DCF}}}& \multicolumn{4}{c|}{\textbf{CAMO-Test}} & \multicolumn{4}{c|}{\textbf{COD10K-Test}} & \multicolumn{4}{c|}{\textbf{NC4K}} \\ \cline{3-14} 
\multicolumn{2}{|c|}{}&\textbf{\textit{$S_\alpha\uparrow$}}& \textbf{\textit{$E_\phi\uparrow$}}& \textbf{\textit{$F^\omega _\beta\uparrow$}}& \textbf{\textit{$\mathcal{M}\downarrow$}}&\textbf{\textit{$S_\alpha\uparrow$}}& \textbf{\textit{$E_\phi\uparrow$}}& \textbf{\textit{$F^\omega _\beta\uparrow$}}& \textbf{\textit{$\mathcal{M}\downarrow$}}&\textbf{\textit{$S_\alpha\uparrow$}}& \textbf{\textit{$E_\phi\uparrow$}}& \textbf{\textit{$F^\omega _\beta\uparrow$}}& \textbf{\textit{$\mathcal{M}\downarrow$}}\\ \hline
\multicolumn{2}{|c|}{baseline}&0.834&0.882&0.750&0.063&0.829&0.890&0.697&0.033&0.854&0.898&0.771&0.045\\ 
\multicolumn{2}{|c|}{w/o DAW}&0.842&0.894&0.768&0.061&0.829&0.886&0.697&0.033&0.858&0.901&0.779&0.044\\ 
\multicolumn{2}{|c|}{w/o CA-SA}&0.842&0.885&0.761&0.061&0.824&0.882&0.685&0.035&0.854&0.895&0.768&0.046\\ \hline
\multicolumn{1}{|c|}{\multirow{4}{*}{DAW}} &First Layer  &0.849&0.897&0.779&0.057&\textbf{0.838}&0.898&0.712&\textbf{0.031}&\textbf{0.865}&\textbf{0.910}&0.790&\textbf{0.041}\\
\multicolumn{1}{|c|}{}&Second Layer &0.856&0.901&0.783&0.056&0.837&0.898&0.709&\textbf{0.031}&0.863&0.906&0.784&0.043\\
\multicolumn{1}{|c|}{}&Third Layer&0.854&0.904&0.789&0.053&0.836&0.898&0.710&\textbf{0.031}&0.863&0.907&0.787&0.042\\
\multicolumn{1}{|c|}{}&ALL Layer&\textbf{0.860}&\textbf{0.913}&\textbf{0.799}&\textbf{0.051}&\textbf{0.838}&\textbf{0.899}&\textbf{0.715}&\textbf{0.031}&\textbf{0.865}&0.909&\textbf{0.792}&0.042\\ \hline
\end{tabular}}
\label{tab:ablation2}
\end{table*}

\begin{figure}[!t]
\centerline{\includegraphics[width=\columnwidth]{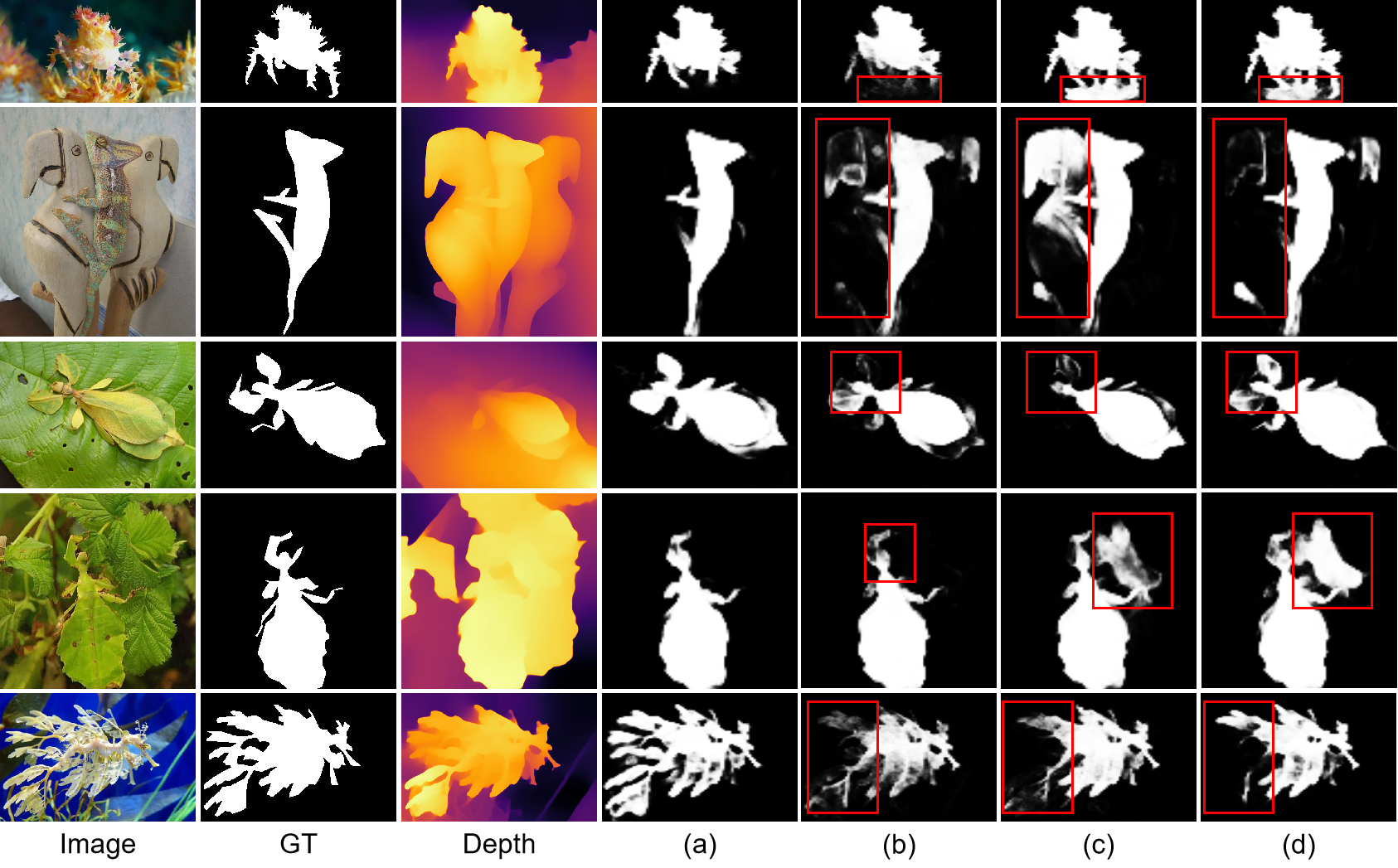}}
\caption{Visual results of the depth-weighted cross-attention fusion module. (a) DCF module fused features; (b) DCF without CA-SA operation fused features; (c) DCF without depth-weighted fused features; (d) baseline fused features. The regions marked by red boxes indicate the differences.}
\label{fig:fig6}
\end{figure}

Besides, we conduct a comparison between DCF and three other feature fusion strategies. The first strategy involves removing the weighting to depth (w/o DAW), the second strategy involves removing the CA-SA operation of DCF module (w/o CA-SA), 
and the third strategy involves removing both the CA-SA operation and the weighting to depth, using only convolution for simple connection (baseline). As shown in Table \ref{tab:ablation2}, the performance of the full DCF significantly outperforms these three fusion strategies. For instance, on the CAMO dataset, the accuracy is improved by 1.8$\%$, 1.8$\%$ and 2.6$\%$ respectively. These results provide empirical evidence for the effectiveness of our proposed cross-attention-based encoder fusion scheme, demonstrating that the weighting of depth information effectively mitigates the negative impact of low-quality depth maps.
For a better reading, we have also provided some qualitative demonstrations in Figure \ref{fig:fig6}.
Observing Figure \ref{fig:fig6} reveals that the model overly relies on the depth map when depth weighting is not employed. This over-reliance leads to inaccurate depth information affecting the accuracy of the detection results. In contrast, our proposed depth-weighted fusion strategy effectively mitigates the effect of inaccurate depth maps, and the attention mechanism we employ aids the model to effectively focus on the details of the target. This strategic approach enables our DAF-Net to dynamically utilize the value of depth information to more accurately focus on camouflaged targets.

To explore the influence of depth maps from different layers on the deep weight representation, we conducted three ablation experiments. Specifically, "First Layer" means utilizing the RGB maps and depth maps from the encoder output of the first layer as inputs for the Deep Attention Weighting module in each layer. Similarly, "Second Layer" denotes employing features from the encoder output of the second layer, at which point the feature fusion of the first layer is not depth-weighted. "Third Layer" means utilizing features from the encoder output of the third layer, whereas the feature fusion of the first two layers is not depth-weighted. Lastly, "All Layer" means weighting the depth of each layer using its respective features. Based on the results shown in Table \ref{tab:ablation2}, we observe that the "All Layer" approach yields the best depth weighting effect and demonstrates the superior effectiveness of the DFC module.

\begin{table*}[!t]
\caption{Ablation study of the architecture of encoder on three datasets. The best results are highlighted in \textbf{Bold}.}
\resizebox{\textwidth}{!}{
\begin{tabular}{|c|cccc|cccc|cccc|}
\hline
\multirow{2}{*}{\textbf{Encoder Architecture}} &\multicolumn{4}{c}{\textbf{CAMO-Test}}&\multicolumn{4}{|c}{\textbf{COD10K-Test}}&\multicolumn{4}{|c|}{\textbf{NC4K}} \\
\cline{2-13} 
&\textbf{\textit{$S_\alpha\uparrow$}}& \textbf{\textit{$E_\phi\uparrow$}}& \textbf{\textit{$F^\omega _\beta\uparrow$}}& \textbf{\textit{$\mathcal{M}\downarrow$}}&\textbf{\textit{$S_\alpha\uparrow$}}& \textbf{\textit{$E_\phi\uparrow$}}& \textbf{\textit{$F^\omega _\beta\uparrow$}}& \textbf{\textit{$\mathcal{M}\downarrow$}}&\textbf{\textit{$S_\alpha\uparrow$}}& \textbf{\textit{$E_\phi\uparrow$}}& \textbf{\textit{$F^\omega _\beta\uparrow$}}& \textbf{\textit{$\mathcal{M}\downarrow$}} \\
\hline
w/o depth&0.797&0.847&0.706&0.078&0.815&0.880&0.673&0.036&0.847&0.893&0.759&0.048 \\
w/o residual addition &0.828&0.878&0.752&0.067&0.822&0.882&0.682&0.035&0.852&0.898&0.767&0.046 \\
w/o ViT      &0.835&0.884&0.757&0.066&0.827&0.887&0.690&0.034&0.854&0.899&0.769&0.046 \\
\hline
DAF-Net&\textbf{0.860}&\textbf{0.913}&\textbf{0.799}&\textbf{0.051}&\textbf{0.838}&\textbf{0.899}&\textbf{0.715}&\textbf{0.031}&\textbf{0.865}&\textbf{0.909}&\textbf{0.792}&\textbf{0.042}\\
\hline
\end{tabular}}
\label{tab:ablation3}
\end{table*}

\begin{table*}[!t]
\caption{Ablation study of Feature aggregation decoder on three datasets. The best results are highlighted in \textbf{Bold}.}
\resizebox{\textwidth}{!}{
\begin{tabular}{|c|cccc|cccc|cccc|}
\hline
\multirow{2}{*}{\centering \textbf{Decoder}}&\multicolumn{4}{|c|}{\textbf{CAMO-Test}}&\multicolumn{4}{|c|}{\textbf{COD10K-Test}}&\multicolumn{4}{|c|}{\textbf{NC4K}} \\
\cline{2-13} 
&\textbf{\textit{$S_\alpha\uparrow$}}& \textbf{\textit{$E_\phi\uparrow$}}& \textbf{\textit{$F^\omega _\beta\uparrow$}}& \textbf{\textit{$\mathcal{M}\downarrow$}}&\textbf{\textit{$S_\alpha\uparrow$}}& \textbf{\textit{$E_\phi\uparrow$}}& \textbf{\textit{$F^\omega _\beta\uparrow$}}& \textbf{\textit{$\mathcal{M}\downarrow$}}&\textbf{\textit{$S_\alpha\uparrow$}}& \textbf{\textit{$E_\phi\uparrow$}}& \textbf{\textit{$F^\omega _\beta\uparrow$}}& \textbf{\textit{$\mathcal{M}\downarrow$}} \\
\hline
w/o G-ECA 
&0.855&0.905&0.786&0.053&0.837&0.895&0.710&0.032&0.864&0.907&0.789&\textbf{0.042} \\
w/o FAM &0.857&0.905&0.792&0.053&\textbf{0.838}&0.898&0.713&0.032&\textbf{0.865}&0.908&0.789&\textbf{0.042} \\
w/o residual addition  &0.854&0.902&0.792&0.052&0.830&0.891&0.699&0.033&0.856&0.902&0.780&0.045 \\ \hline
DAF-Net		     &\textbf{0.860}&\textbf{0.913}&\textbf{0.799}&\textbf{0.051}&\textbf{0.838}&\textbf{0.899}&\textbf{0.715}&\textbf{0.031}&\textbf{0.865}&\textbf{0.909}&\textbf{0.792}&\textbf{0.042}\\
\hline
\end{tabular}}
\label{tab:ablation5}
\end{table*}

\noindent{\textbf{Effectiveness of Three-branch architecture.}}
The fusion map contains a substantial amount of valuable feature information. To fully capture the contextual information and ensure comprehensive extraction and utilization of these fusion features, we employ the R50+ViT-B/16\cite{dosovitskiy2020image} as our fusion subnetwork. To validate the effectiveness of this subnetwork, we conduct a series of ablation experiments, the results of which are presented in Table \ref{tab:ablation3}.

Firstly, we remove the deep branch and fusion subnetwork, transforming the network into a basic single-branch architecture with only RGB maps as input. The experimental results are depicted in the first row of Table \ref{tab:ablation3} (w/o depth). 
Subsequently, we remove the entire fusion subnetwork and only employ convolutional operations to resize each stage of the fusion map. The corresponding experimental results are shown in the second row of Table \ref{tab:ablation3} (w/o residual addition). 
Following this, we reintroduce the residual module in the fused subnetwork while keeping ViT stripped, enabling us to assess the extraction effect of the Transformer on fused features. The experimental results are presented in the third row of Table \ref{tab:ablation3} (w/o ViT). 
The efficacy of depth information is validated through a comparison of quantization results between the first row and the subsequent two rows.
Furthermore, by comparing the quantitative results of the second and third rows, we confirm the complementary effect of residuals on the fused features. Comparing the quantitative results of the third and fourth rows, we observe that the ViT architecture can more effectively capture the fused information.

\noindent{\textbf{Effectiveness of RMFE.}}
%To increase the confidence in the more precise RGB information, we incorporate the enhanced RGB feature residuals. The results from \#4 and \#5 in Table \ref{tab:ablation2} demonstrate that the enhanced RGB information improves the network's performance on the two datasets to different extents.
We conduct ablation experiments to assess the effectiveness of the feature enhancement module RMFE\cite{zhu2022can}.
The experimental results are presented in Table \ref{tab:ablationall}.

To enhance the receptive field of fused features and increase the confidence in the more precise RGB information, we conduct feature enhancement on both RGB and fused features using RMFE. Additionally, we employ residual addition to further amplify the contribution of RGB features in generating the final prediction map. Specifically, as detailed in Table \ref{tab:ablationall}, we contrast the network with RMFE removed (denoted as \#4) against the network with RMFE included (denoted as \#5). The results in Table \ref{tab:ablationall} demonstrate that both enhanced fused features and RGB features improve the network's performance on the two datasets to different extents.
% To increase the confidence in the more precise RGB information, we incorporate the enhanced RGB feature residuals. Specifically, as detailed in Table \ref{tab:ablationall},  we compared a network that removed the decoder output features and augmented RGB features for residual summing (referred to as \#4) with one that included the enhanced RGB feature residual (referred to as \#5). 
% The results from \#4 and \#5 in Table \ref{tab:ablationall} demonstrate that the enhanced RGB information improves the network's performance on the two datasets to different extents.

\noindent{\textbf{Effectiveness of the Feature Aggregation Decoder.}}
We perform ablation experiments on our proposed Feature Aggregation Decoder to showcase its effectiveness. The results of these experiments are illustrated in Tables \ref{tab:ablationall} and \ref{tab:ablation5}.

We start by removing all critical modules, using a three-branch architecture and base fusion modules for fusion, and adopting simple concat and convolution as decoders.  As shown in Table \ref{tab:ablationall} the baseline version (denoted as \#1) yields the lowest scores across all evaluation metrics. Building upon this baseline, we enhance the model by replacing the base Decoder with our designed Feature Aggregation Decoder (denoted as \#2).  Comparing \#2 with the baseline, we observe improved performance in all evaluation metrics. Specifically, the four indicators show average improvements of 2.1$\%$, 3.4$\%$, 1.15$\%$ and 0.55$\%$. These experimental results affirm that our Feature Aggregation Decoder significantly contributes to more accurate detection of camouflaged objects within the network.

Furthermore, we evaluate the detection performance of each component in the Feature Aggregation Decoder individually. Table \ref{tab:ablation5} presents the results when removing the vanilla ECA (w/o G-ECA), the Feature Aggregation Module (w/o FAM) and the residual addition between the decoder and the three RGB features (w/o residual addition). The quantitative comparison demonstrates that each module and design contributes to improve the overall detection performance of the decoder.

\noindent{\textbf{Res2Net vs ResNet50.}}
In order to account for the distinctive properties of the depth and image branches, we employ distinct backbones for feature extraction. 
ResNet50's deep and residual techniques equip it with the capability to acquire more intricate features, making it the preferred backbone network for feature extraction in the majority of existing COD models. Nevertheless, we contend that Res2Net's utilization of multi-scale convolution, as opposed to ResNet50, enhances its ability to extract multi-scale features, consequently boosting model detection accuracy. 
Table \ref{tab:ablationall} illustrates the efficacy of employing either ResNet50 or Res2Net for the depth and RGB branches, with both backbones yielding favorable outcomes. Specifically, configuration \#7 stands out as the optimal choice, striking a balance between performance and parameter efficiency.

%After careful evaluation of the tradeoff between performance and the number of parameters, we find that \#7 in Table \ref{tab:ablation2} yields the optimal result and was selected in our proposed model.

\subsection{Limitation}
\begin{figure}[b]
\centerline{\includegraphics[width=\columnwidth]{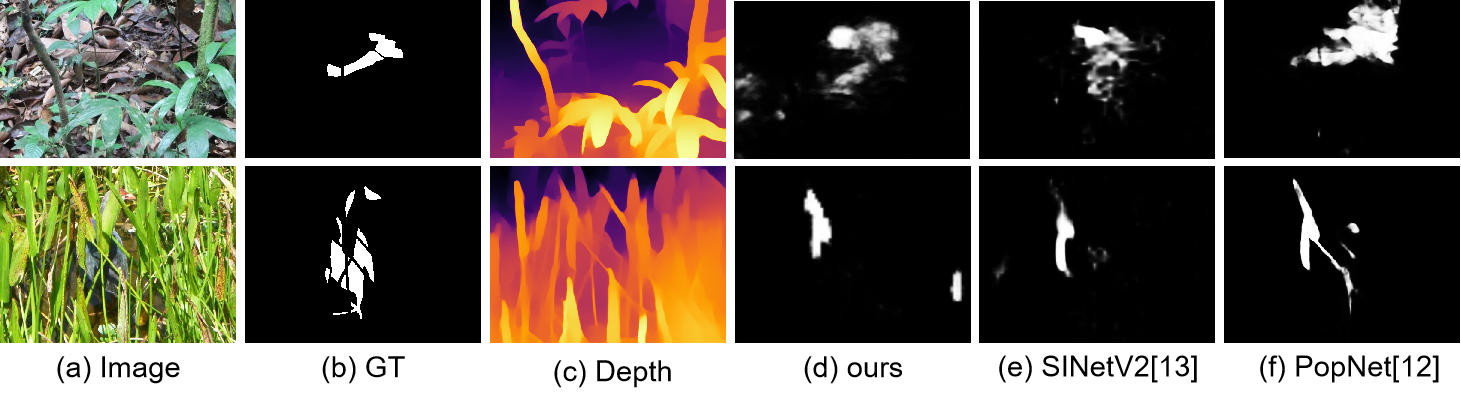}}
\caption{Failure cases of our DAF-Net.}
\label{fig:fig7}
\end{figure}

Despite the positive outcomes demonstrated by various experiments regarding the performance of the method presented in this paper, instances of failure persist, as exemplified in Figure \ref{fig:fig7}.
In complex environments, factors such as the target's distance from the camera, occlusion by numerous unrelated objects and the presence of shadows can lead to misleading depth information about non-camouflaged objects for the detection of camouflaged objects.
Unfortunately, the methods outlined in this paper are not fully effective in addressing these challenges. It's important to emphasize that this issue persists even within the domain of salient object detection, which leverages depth maps acquired from depth sensors for its enhancement \cite{zhou2021rgb}.

Furthermore, achieving perfect segmentation of camouflaged targets from their surroundings has been a difficult challenge in computer vision. The SIDE network fundamentally differs from the camouflaged object detection network specifically designed for these camouflaged objects, leading to occasional difficulties in accurately estimating the true depth of camouflaged objects concealed within complex surroundings. How to accurately detect camouflaged targets in scenarios with poor quality depth maps and how to obtain more accurate depth maps are still open areas to be explored.

\section{DISCUSSIONS AND CONCLUSION}
In this paper, we introduce the DAF-Net which explores the use of scene depth predicted by depth estimators for camouflaged object detection. Different from sensored depth, the generated depth contains noisy information, which should be selectively utilized.

Accurately detecting camouflaged objects with depth maps of uncertain quality is the focus of this paper. We explore the approach of adaptively weighting the depth information to address this challenge. 
In contrast to prior networks, our approach employs a three-branch encoder to extract RGB, depth and fusion information separately. Within this process, we employ our proposed Depth-weighted Cross-attention Fusion (DCF) module to adaptively fuse the depth and RGB information. Subsequently, we utilize our proposed Feature Aggregation Decoder (FAD) to fuse the enhanced aggregated features. We demonstrate the superiority of our method by performing qualitative and quantitative experiments on three public datasets.

The broader impact of this paper is to demonstrate the possibility of directly utilizing the generated depth map as input in the COD domain. Our work set up a bridge between SIDE and COD, where COD can benefit from the generated depth map to some extent. Furthermore, our fusion approach has the potential to provide inspiration for other research efforts focused on multimodal fusion techniques. Our approach can be extended to the SOD domain as well as to more realistic scenarios of camouflaged target detection.

Despite the inclusion of depth map auxiliary in our approach,
we still face difficulties such as information redundancy and depth inaccuracy in complex environments. In forthcoming research, we will contemplate improving the image resolution of the input depth estimation network, employing image preprocessing techniques to emphasize the camouflaged object and enhance the quality of the generated depth map. Grafting depth estimation network with COD network is another potential direction. Additionally, collecting COD datasets with sensored depth although requires much effort but may boost the performance according to the findings in this paper.

\bibliographystyle{cas}

\bibliography{cas}

% % Here goes the biography details.

% % Here goes the biography details.

\end{document}